\documentclass{article}

\usepackage{microtype}
\usepackage{graphicx}
\usepackage{subfigure}
\usepackage{booktabs} 
\usepackage{amsmath}
\usepackage{amsfonts}
\usepackage{hyperref}

\usepackage[accepted]{icml2019}

\icmltitlerunning{Compositional Structure Learning for Sequential Video Data}

\begin{document}

\twocolumn[
\icmltitle{Compositional Structure Learning for Sequential Video Data}




\begin{icmlauthorlist}
\icmlauthor{Kyoung-Woon On}{to}
\icmlauthor{Eun-Sol Kim}{goo}
\icmlauthor{Yu-Jung Heo}{to}
\icmlauthor{Byoung-Tak Zhang}{to}

\end{icmlauthorlist}

\icmlaffiliation{to}{Department of Computer Science and Engineering, Seoul National University, Seoul, South Korea}
\icmlaffiliation{goo}{Kakao Brain, Seongnam, South Korea}

\icmlcorrespondingauthor{Byoung-Tak Zhang}{btzhang@snu.ac.kr}

\icmlkeywords{Machine Learning, ICML}

\vskip 0.3in
]



\printAffiliationsAndNotice{}  

\begin{abstract}
Conventional sequential learning methods such as Recurrent Neural Networks (RNNs) focus on interactions between consecutive inputs, i.e. first-order Markovian dependency. 
However, most of sequential data, as seen with videos, have complex temporal dependencies that imply variable-length semantic flows and their compositions, and those are hard to be captured by conventional methods.
Here, we propose Temporal Dependency Networks (TDNs) for learning video data by discovering these complex structures of the videos.
The TDNs represent video as a graph whose nodes and edges correspond to frames of the video and their dependencies respectively. 
Via a parameterized kernel with graph-cut and graph convolutions, the TDNs find compositional temporal dependencies of the data in multilevel graph forms.
We evaluate the proposed method on the large-scale video dataset Youtube-8M.
The experimental results show that our model efficiently learns the complex semantic structure of video data.
\end{abstract}

\section{Introduction}
A fundamental problem in learning sequential data is to find semantic structures underlying the sequences for better representation learning.
In particular, the most challenging problems are to segment the whole long-length sequence in multiple semantic units and to find their compositional structures.

In terms of neural network architectures, many problems with sequential inputs are resolved by using Recurrent Neural Networks (RNNs) as it naturally takes sequential inputs frame by frame. However, as the RNN-based methods take frames in (incremental) order, the parameters of the methods are trained to capture patterns in transitions between successive frames, which makes it hard to find long-term temporal dependencies through overall frames. For this reason, their variants, such as Long Short-Term Memory (LSTM)~\cite{hochreiter1997long} and Gated Recurrent Units (GRU)~\cite{chung2014empirical}, have made the suggestion of ignoring noisy (unnecessary) frames and maintaining the semantic flow through the whole sequence by turning switches on and off. However, it is still hard to retain multiple semantic flows and to learn their hierarchical and compositional relationships.

In this work, we propose Temporal Dependency Networks (TDNs) which can discover composite dependency structure in video inputs and utilize them for representation learning of videos. 
The composite structures are defined as a multilevel graph form, which make it possible to find long-length dependencies and its hierarchical relationships effectively.

A single video data input is represented as a temporal graph, where nodes and edges represent frames of the video and relationships between two nodes.
From the input representations, the TDNs find composite temporal structures in the graphs with two key operations: temporally constrained normalized graph-cut and graph convolutions.
A set of semantic units is found by cutting the input graphs with temporally constrained normalized graph-cuts.
Here, the cutting operator is conducted with the weighted adjacency matrix of the graph which is estimated by parameterized kernels. 
After getting a new adjacency matrix with the cutting operations, representations of the inputs are updated by applying graph convolutional operations.
As a result of stacking those operations, compositional structures between whole frames are discovered in a multilevel graph form.
Furthermore, the proposed method can be learned in an end-to-end manner.

We evaluate our method with the YouTube-8M dataset, which is for the video understanding task. 
As a qualitative analysis of the proposed model, we visualize semantic temporal dependencies of sequential input frames, which are automatically constructed. 


The remainder of the paper is organized as follows. To make further discussion clear, the problem statement of this paper are described in the following sections. After that, Temporal Dependency Networks (TDNs) are suggested in detail and the experimental results with the real dataset YouTube-8M are presented. 

\begin{figure*}[ht]
\centering

\includegraphics[width=0.9\textwidth]{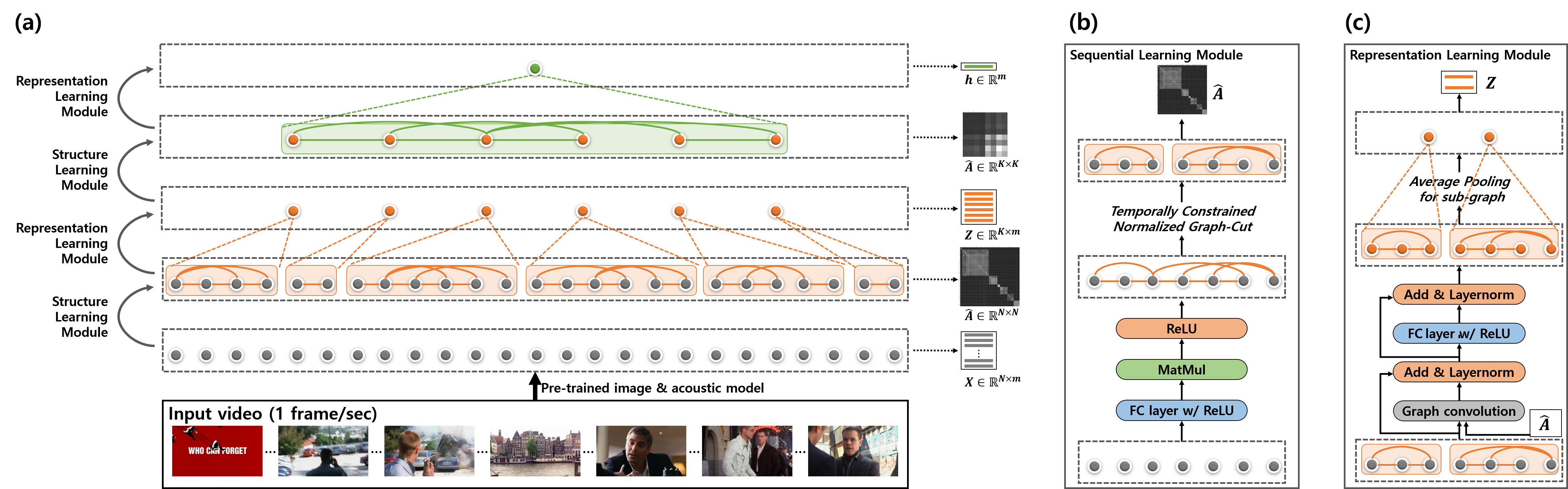}

\caption{(a): Overall architecture of the Temporal Dependency Networks (TDNs) for a video classification task. (b), (c): sophisticated illustrations of Structure Learning Module and Representation Learning Module.
}
\label{fig:model_figure}
\vspace{-0.5cm}
\end{figure*}

\section{Problem Statement}

We consider videos as inputs, and a video is represented as a graph $G$.
The graph $G$ has nodes corresponding to each frame in the video with feature vectors
and the dependencies between two nodes are represented with weight values of corresponding edges.

Suppose that a video $X$ has $N$ successive frames 
and each frame has an $m$-dimensional feature vector $x\in \mathbb{R}^{m}$.
Each frame corresponds to a node $v \in V$ of graph $G$, and the dependency between two frames $v_i$, $v_j$ is represented by a weighted edge $e_{ij} \in E$.
From $G=(V,E)$, the dependency structures among video frames is defined as the weighted adjacency matrix $A$, where $A_{ij}=e_{ij}$.
With aforementioned notations and definitions, we can now formally define the problem of video representations learning as follows:


\textit{Given the video frames representations $X\in \mathbb{R}^{N\times m}$, we seek to discover a weighted adjacency matrix $A\in \mathbb{R}^{N\times N}$ which represents dependency among frames.}
\begin{equation}
  f:X\rightarrow A 
\end{equation}
\textit{With $X$ and $A$, final representations for video  $h\in \mathbb{R}^{l}$ are acquired by $g$.}
\begin{equation}
  g:\{X,A\}\rightarrow h 
\end{equation}
The obtained video representations $h$ can be used for various tasks of video understanding. In this paper, the multi-label classification problem for video understanding is mainly considered.

\section{Temporal Dependency Networks}

The Temporal Dependency Networks (TDNs) consist of two modules: a structure learning module with the graph-cuts and a representation learning module with graph convolutions.

In the structure learning module, the dependencies between frames $\hat{A}$ are estimated via parameterized kernels and the temporally constrained graph-cut algorithm.
The suggested graph-cut algorithm i) makes the dense dependencies to be sparse, and ii) forms a set of temporally non-overlapped semantic units (disjoint sub-graphs) to construct the compositional hierarchies.

After getting the matrix $\hat{A}$, representations of the inputs are updated by applying graph convolutions followed by pooling operations.

As mentioned earlier in this work, furthermore, by stacking those modules, compositional structures of whole frames are discovered in a multilevel graph form.
Figure \ref{fig:model_figure}(a) illustrates the whole structure of TDNs.
In the next sections, operations of each of these modules are described in detail.

\subsection{Structure Learning Module} \label{sec:SLL}
The structure learning module is composed of two steps. The first step is to learn the dependencies over all frames via the parameterized kernel $\mathcal{K}$:
\begin{equation}\label{eq:kernel}
  \hat{A}_{ij} = \mathcal{K}(x_i,x_j) = ReLU(f(x_i)^\top f(x_j))
\end{equation}
where $f(x)$ is a single-layer feed-forward network without non-linear activation:
\begin{equation}
  f(x) = W^fx+b^f
\end{equation}
with $W^f\in\mathbb{R}^{m\times m}$ and $b^f\in\mathbb{R}^{m}$.

Then, the $\hat{A}$ is refined by the normalized graph-cut algorithm~\cite{shi2000normalized}.
The objective of the normalized graph-cut is:
\begin{equation}
  Ncut(V_1,V_2) = {\sum_{v_i\in V_1, v_j\in V_2} \hat{A}_{ij} \over \sum_{v_i\in V_1} \hat{A}_{i\cdot }} + {\sum_{v_i\in V_1, v_j\in V_2} \hat{A}_{ij} \over \sum_{v_j\in V_2} \hat{A}_{j\cdot}}
\end{equation}
It is formulated as a discrete optimization problem and usually relaxed to continuous, which can be solved by eigenvalue problem with the $O(n^2)$ time complexity~\cite{shi2000normalized}.
The video data is composed of time continuous sub-sequences so that no two partitioned sub-graphs have an overlap in physical time. Therefore, we add the temporal constraint~\cite{rasheed2005detection,sakarya2008graph} as follows,
\begin{equation}\label{eq:NCut_Const}
  (i<j\; \text{or}\; i>j)\;\; \text{for all}\;v_{i}\in V_1, v_j\in V_2
\end{equation}
Thus, a cut can only be made along the temporal axis and complexity of the graph partitioning is reduced to linear time.
We apply the graph-cut recursively so that the refined $\hat{A}$ and multiple partitioned sub-graphs are obtained.
The number of sub-graph $K$ is determined by the length of the video $N$. 
\begin{equation}\label{eq:threshold}
  K = 2^{\lfloor\log_{2}\sqrt{N}\rfloor-1}
\end{equation}
Figure \ref{fig:model_figure}(b) depicts the detailed  operations of the structure learning module.

\subsection{Representation Learning Module}
After estimating the weighted adjacency matrix $\hat{A}$, the representation learning module updates the representations of each frame via a graph convolution operation~\cite{kipf2016semi} followed by a position-wise fully connected network. We also employ a residual connection~\cite{he2016deep} around each layer followed by layer normalization~\cite{ba2016layer}:

\begin{align}
  Z' &= LN(\sigma(\hat{D}^{-{1}}\hat{A}XW^{Z'})+X) \\
  Z &= LN(\sigma(Z'W^{Z}+b^{Z})+Z')
\end{align}
where $W^{Z'},W^{Z}\in\mathbb{R}^{m\times m}$ and $b^{Z}\in\mathbb{R}^{m}$.

Once the representations of each frame are updated, an average pooling operation for each partitioned sub-graph is applied.
Then we can obtain higher level representations $Z\in\mathbb{R}^{K\times m}$, where $K$ is the number of partitioned sub-graphs (Figure \ref{fig:model_figure}(c)). 
In the same way, $Z$ is fed into the new structure learning module and we can get the the  video-level representation $h\in \mathbb{R}^m$.
Finally, labels of the video can be predicted using a simple classifier.

\section{Experiments}
\subsection{YouTube-8M Dataset}

YouTube-8M~\cite{abu2016youtube} is a benchmark dataset for video understanding, where the task is to determine the key topical themes of a video.
It consists of 6.1M video clips collected from YouTube and the video inputs consist of two multimodal sequences, which are the image and audio. 
Each video is labeled with one or multiple tags referring to the main topic of the video. 
The dataset split into three partitions, 70\% for training, 20\% for validation and 10\% for test. As we have no access to the test labels, all results in this paper are reported for validation set.

Each video is encoded at 1 frame-per-second up to the first 300 seconds.
As the raw video data is too huge to be treated, each modality is pre-processed with pretrained models by the author of the dataset.
More specifically, the frame-level visual features were extracted by inception-v3 network~\cite{szegedy2016rethinking} trained on ImageNet and the audio features were extracted by VGG-inspired architecture~\cite{hershey2017cnn} trained for audio classification. 
PCA and whitening method are then applied to reduce the dimensions to 1024 for the visual and 128 for audio features.

Global Average Precision (GAP) is used for the evaluation metric for the multi-label classification task as used in the YouTube-8M competition. For each video, 20 labels are predicted with confidence scores. Then the GAP score computes the average precision and recall across all of the predictions and all the videos.


\begin{figure*}[!ht]
\begin{center}
\includegraphics[width=0.85\textwidth]{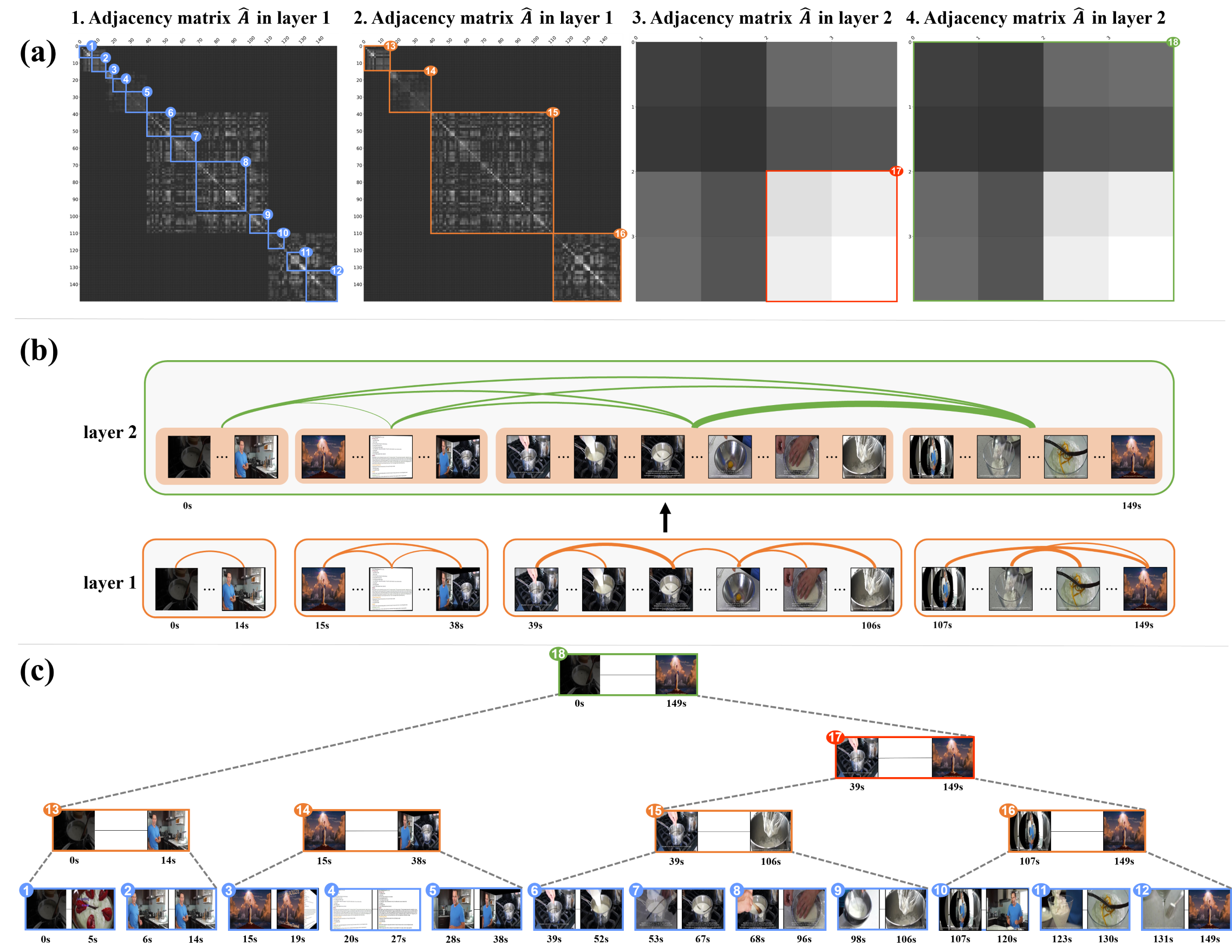}

\caption{An example of the constructed temporal dependency structures for a real input video, titled "Rice Pudding" (https://youtu.be/cD3enxnS-JY) is visualized. The topical themes (labels) of this video are \{Food, Recipe, Cooking, Dish, Dessert, Cake, baking, Cream, Milk, Pudding and Risotto\}. (a): Learned adjacency matrices in the layer 1 and 2 are visualized. The strength of connection are encoded in a gray-scale which 1 to white and 0 to black. (a)-1: 12 bright blocks in layer 1 are detected (blue rectangles), each block (highly connected frames) represents a semantic unit. (a)-2: Sub-graphs of the input are denoted by orange rectangles. It shows that the semantically meaningful scenes are found by temporally constrained graph-cut.  (a)-3 and (a)-4: learned high-level dependency structures in layer 2 are revealed with red and green rectangles. (b): The conceptual illustration of learned temporal dependency is shown. In the 1st layer, the temporal dependency structure is learned only within the sub-graphs. In the 2nd layer, inter connections of sub-graphs are learned to capture high-level temporal dependencies. (c): The whole composite temporal dependencies are presented.} 
\vspace{-0.5cm}
\label{fig:qualitative_result}
\end{center}
\end{figure*}

\subsection{Qualitative results: Learning compositional temporal dependencies}\label{sec:qr}

In this section, we demonstrate compositional learning capability of TDNs by analyzing constructed multilevel graphs.
To make further discussion clear, four terms are used to describe the compositional dependency structure in input video: semantic units, scenes, sequences and a video for each level.
In Figure \ref{fig:qualitative_result}, a real example with the usage of video titled ``Rice Pudding\footnote{https://youtu.be/cD3enxnS-JY}'' is described to show the results. 

In Figure \ref{fig:qualitative_result}(a), the learned adjacency matrices in each layer are visualized in gray-scale images: two of the left are obtained from the 1st layer and two of the right from the 2nd layer. To denote multilevel semantic flows, four color-coded rectangles (blue, orange, red and green) are marked and those colors are consistent with Figure \ref{fig:qualitative_result}(b) and (c).

Along with diagonal elements of the adjacency matrix in the 1st layer (Figure \ref{fig:qualitative_result}(a)-1), a set of semantic units are detected corresponding to bright blocks (blue). 
Interestingly, we could find that each semantic unit contains highly correlated frames.
For example, the \#1 and \#2 are each shots introducing the YouTube cooking channel and how to make rice pudding, respectively. The \#4 and \#5 are shots showing a recipe of rice pudding and explaining about the various kinds of rice pudding. The \#6 and \#7 are shots putting ingredients into boiling water in the pot and bringing milk to boil along with other ingredients. At the end of the video clip, \#11 is a shot decorating cooked rice pudding and \#12 is an outro shot that invites the viewers to subscribe.

These semantic units compose variable-length scenes of the video, and each scene corresponds to a sub-graph obtained via graph-cut (Figure \ref{fig:qualitative_result}(a)-2.). 
For example, \#13 is a scene introducing this cooking channel and rice pudding. Also, \#15 is a scene of making rice pudding step by step with details and \#16 is an outro scene wrapping up with cooked rice pudding. 
The 1st-layer of the model updates representations of frame-level nodes with these dependency structures, then aggregates frame-level nodes to form scene-level nodes (Layer 1 in the Figure \ref{fig:qualitative_result}(b)).

In Figure \ref{fig:qualitative_result}(a)-3 and (a)-4, the sequence-level semantic dependencies (red) are shown. The \#17 denotes a sequence of making rice pudding from beginning to end, which contains much of the information for identifying the topical theme of this video. Finally, the representations of scenes are updated and aggregated to get representations of the whole video (Layer 2 in the Figure \ref{fig:qualitative_result}(b)). The Figure \ref{fig:qualitative_result}(c) presents the whole composite temporal dependencies.



\section{Conclusions}
In this paper, we proposed TDNs which learn not only the representations of multimodal sequences, but also composite temporal dependency structures within the sequence. The qualitative experiment is conducted on a real large-scale video dataset and shows that the proposed model efficiently learns the inherent dependency structure of temporal data. 

\section*{Acknowledgements}
This work was partly supported by the Korea government (2015-0-00310-SW.StarLab, 2017-0-01772-VTT, 2018-0-00622-RMI, 2019-0-01367-BabyMind, 10060086-RISF, P0006720-GENKO).


\bibliography{example_paper}
\bibliographystyle{icml2019}

\end{document}